\documentclass[11pt]{article}
\usepackage[a4paper,margin=1in]{geometry}
\usepackage[T1]{fontenc}
\usepackage[utf8]{inputenc}
\usepackage{lmodern}
\usepackage{amsmath,amssymb}
\usepackage{graphicx}
\usepackage{booktabs}
\usepackage{array}
\usepackage{microtype}
\usepackage{xurl}
\usepackage{hyperref}
\usepackage{enumitem}
\usepackage{placeins}
\hypersetup{colorlinks=true,allcolors=blue}
\graphicspath{{figures/}}

\title{ScienceBuddy-PV: A Verifier-Governed Recursive Improvement Protocol for Panvascular Intervention Research}
\author{\parbox{0.96\textwidth}{\centering
\normalsize Lingsen You$^{1,2,*}$ \quad Yujun Guo$^{3,*}$ \quad Xinyu Zhong$^{4,*}$ \quad Zijie Feng$^{5,*}$\\[3pt]
\normalsize Wentong Wang$^{1,2,*}$ \quad Li Shen$^{1,2,\dagger}$ \quad Junbo Ge$^{1,2,\dagger}$\\[7pt]
\small $^{1}$Department of Cardiology, Zhongshan Hospital, Fudan University, Shanghai Institute of Cardiovascular Diseases, State Key Laboratory of Cardiovascular Diseases, NHC Key Laboratory of Ischemic Heart Diseases China, National Clinical Research Center for Interventional Medicine, Key Laboratory of Viral Heart Diseases, Chinese Academy of Medical Sciences, Shanghai 200032, China.\\[2pt]
\small $^{2}$Institutes of Biomedical Sciences, Fudan University, Shanghai 200030, China.\\[2pt]
\small $^{3}$College of Biomedical Engineering, Fudan University, Shanghai 200438, China.\\[2pt]
\small $^{4}$Department of Cardiology, Shidong Hospital, Yangpu District, Shanghai 200438, China.\\[2pt]
\small $^{5}$Department of Chemical and Environmental Engineering, University of Nottingham, University Park, Nottingham NG7 2RD, United Kingdom.\\[2pt]
\small $^{*}$Joint first authors; $^{\dagger}$Corresponding authors.\\[2pt]
\small Correspondence: shen\_li@fudan.edu.cn; jbge@zs-hospital.sh.cn}}
\date{9 October 2026}

\begin{document}
\maketitle

\begin{abstract}
\textbf{Background:} Panvascular intervention research must connect heterogeneous evidence, vessel geometry, device properties, blood flow, biology and time. A language-model agent can make this work faster, but a plausible answer is not the same as a traceable scientific result. ScienceBuddy proposes a recursive-in-recursive loop in which a fixed model is used to improve its harness, and a fixed harness is then used to improve the model from rubric-scored trajectories.
\textbf{Objective:} We adapt this design to panvascular research and audit a fixed verifier component on a public coronary angiography benchmark.
\textbf{Methods:} ScienceBuddy-PV specifies a structured task and evidence ledger, a four-axis device--vessel ontology, an explicit abstention contract and a retained failure bank. The public component experiment uses the ARCADE syntax and stenosis environments, with 1,000 training, 200 validation and 300 test images per task. Numeric identifiers were audited before modelling and no cross-task label was attached to an unrelated frame. A shared task-masked convolutional model was trained once, then held fixed while a heuristic evidence gate and action-state audit were evaluated; the stenosis score threshold was selected on validation data. Recursive harness optimization and the outer model-learning loop remain protocols rather than executed experiments in this study.
\textbf{Results:} On the locked test environments, syntax Dice was 0.475 (95\% CI 0.459--0.490) and stenosis Dice was 0.231 (0.214--0.249); stenosis AUROC and AUPRC were 0.957 and 0.180 at a pixel prevalence of 0.00975. The evidence gate accepted 76.7\% of stenosis-frame outputs versus 100\% for a score-only comparator. After controlled horizontal reflection, vertical shift and intensity attenuation, stenosis Dice was 0.195, 0.228 and 0.170, respectively. Constructed-fault acceptance was 2.0\% for reflection, 6.0\% for shift, 3.0\% for random-frame candidate-mask substitution and 11.7\% for dilation. Reference candidate masks were accepted in 44.3\% of cases and rejected in 55.7\%.
\textbf{Conclusions:} A ScienceBuddy-inspired panvascular harness can make support, abstention and distribution-shift failure measurable on public data. The result is an image-level technical validation and a reproducible protocol for future evidence-agent and device--vessel studies; it is not a patient-level digital twin, a treatment-effect estimate or a clinical device recommendation.
\end{abstract}

\section{Introduction}

Panvascular intervention treats a connected vascular system rather than an isolated image. Device performance depends on geometry, wall mechanics, flow, inflammation, healing and the time course of the intervention. This motivates the device--vessel suitcordance perspective, which separates short-term suitability from long-term concordance and organizes the problem across arterial beds \cite{you_suitcordance,you_radar}. The public Suitcordance Bench makes the idea concrete with geometric, mechanical, haemodynamic and biological axes, while explicitly describing itself as a research prototype rather than a clinical decision tool \cite{suitbench}. The FLOW framework, flexible-electronics work and cardiovascular digital-twin reviews provide complementary experimental and computational layers \cite{flow,4a,digitaltwin}.

Prior work on AI-assisted coronary planning and IVUS segmentation illustrates the value of image-derived procedural support, while also motivating a clear separation between measured anatomy and device-choice claims \cite{samant,ivus}. The scientific bottleneck is therefore procedural as well as statistical. A useful agent must find the right source, preserve the exact endpoint and time window, distinguish a patient from a lesion, connect a proposed action to a supported vessel state, run an analysis that can be reproduced, and state when evidence is insufficient. Generic retrieval-augmented generation does not guarantee any of these properties. In a panvascular setting, a confident error can arise from endpoint conflation, a wrong denominator, an unverified device specification or an invalid transfer of a coronary rate to a peripheral or cerebrovascular bed.

ScienceBuddy is a useful systems template for this problem. Its public preprint turns researcher requests, corrections and execution evidence into executable tasks and evaluation rubrics. The inner recursion holds the task model fixed while a helper proposes bounded changes to instructions, tools, skills or context; candidates are retained only after paired evaluation. The outer recursion keeps the selected harness fixed while fresh on-policy attempts are scored by task rubrics and used for model learning \cite{sciencebuddy}. The preprint reports scientific-agent benchmark cases, not clinical validation, and its harness experiment used simulated feedback. We therefore treat the architecture as a workflow design principle and evaluate one safety-critical component on a public vascular environment.

Related work informs the evaluation contract rather than establishing a world model in the present study. JEPA-Anything uses predictive factorization across heterogeneous domains, motivating explicit state representations and matched representation comparisons \cite{jepa}. The Immune World Model makes interventions explicit and freezes the selected model before independent confirmation, illustrating the separation of hypothesis generation from confirmation \cite{iwm}. We adopt these evaluation principles in the protocol and expose structured mask states and deterministic audit actions. Our projection penalty is only a representation probe: the present model learns neither action-conditioned state transitions nor a validated vascular world model, and its test audit is not independent experimental confirmation.

ScienceIDE offers a complementary approach to making scientific codebases executable learning environments \cite{scienceide}; the proposed task packages below follow this reproducibility goal without claiming that such an agent environment was benchmarked here.

We introduce ScienceBuddy-PV, a domain-specific research protocol with four contributions:

\begin{enumerate}[leftmargin=*]
\item a task and evidence ontology that makes vascular bed, device, endpoint, denominator and time window explicit;
\item a specified verifier-governed inner loop that separates predictive scores from support for an intervention anchor and includes an abstention state;
\item a public-data benchmark with locked test evaluation, controlled shifts, fault injection and an honest record of what is not identifiable;
\item an outer-loop specification for researcher-rubric learning, with public-data and synthetic extensions that can be executed without exposing patient information.
\end{enumerate}

The central claim is deliberately narrow: a scientific agent for panvascular research should publish its evidence contract and failure behavior alongside its predictive metric. The present study tests that claim at image level.

\section{System design}

ScienceBuddy-PV specifies four layers. The proposed workspace stores source files, code, model versions and execution traces; the evidence ledger stores source snapshots and structured evidence units. The scientific tool layer is designed to call literature retrieval, image processing, statistics, CFD/FEM and FLOW-style experimental modules. The improvement layer defines a harness loop and a model-learning loop. The public experiment implements the image model, support gate and audit components. It does not implement or evaluate the full evidence ledger, researcher feedback cycle or recursive harness optimization.

\subsection{Task and evidence contracts}

Every research unit is a \texttt{TaskSpec}, not an unstructured chat turn:

\begin{verbatim}
task_id, cutoff_date, vascular_bed, lesion, device, comparator,
estimand, endpoint_construct, trigger, horizon, denominator,
allowed_sources, required_outputs, abstention_trigger
\end{verbatim}

Every extracted claim is an \texttt{EvidenceItem} with a source URL or DOI, a content hash, an exact locator, the population, vessel bed, device, comparator, endpoint, time window, analysis unit, estimate and uncertainty. A claim inferred from a simulation or transferred across vascular beds is marked as \texttt{inference}; it cannot be written into an observed-effect field. This schema is designed to catch the common error of treating primary patency, restenosis, target-lesion revascularization, major adverse limb events and mortality as interchangeable outcomes.

\subsection{Panvascular ontology}

The ontology keeps four Suitcordance axes separate until each has a defensible label:

\begin{itemize}[leftmargin=*]
\item \textbf{Geometry (G):} diameter, length, taper, branch topology, lesion coverage, calcification and residual stenosis.
\item \textbf{Mechanics (M):} compliance mismatch, over-expansion, wall strain, fatigue and device--wall interface.
\item \textbf{Haemodynamics (H):} lumen recovery, pressure and flow, wall shear, branch disturbance and distal runoff.
\item \textbf{Biology (B):} inflammation, endothelial healing, neointimal drive, drug action and degradation.
\end{itemize}

A composite score may be formed only on an independent calibration set and only when all required axes are observed or the missingness is explicitly reported. Device price, contrast volume, procedure time, bleeding, embolization and other clinical costs are additional decision dimensions; they are not silently absorbed into a Suitcordance score.

\subsection{Recursive improvement protocol}

In the \emph{inner loop}, the task model is fixed. The helper reads execution traces, failed evidence matches and reviewer feedback, and proposes bounded edits to retrieval queries, extraction templates, unit checks, topology checks, statistical validators, image-quality gates and refusal rules. A candidate harness runs on the same development tasks, seeds and compute budget as its parent. It is accepted only if it is executable, improves the paired development score and does not increase a prespecified safety-error rate. The evaluation rubric and an independent red-team set remain frozen.

In the \emph{outer loop}, the selected harness is fixed. New tasks are sampled from researcher requests, public evidence updates and controlled environment perturbations. A rubric combines executable checks with human review of evidence entailment, endpoint consistency, statistical validity, reproducibility, uncertainty and safety. Fresh on-policy trajectories may then support DPO, LoRA or GRPO updates. The updated model is re-evaluated with the inherited harness and the frozen test set before deployment. Correct abstention and escalation to a human are positive outcomes. Citation quantity, answer length and unsupported device ranking are not.

The present paper reports a fixed-verifier component benchmark that can support a future inner loop. It does not report recursive harness optimization or an executed outer model-learning loop.

\section{Public benchmark and methods}

\subsection{ARCADE data and provenance}

We used the public ARCADE coronary x-ray angiography release and its separate syntax and stenosis environments \cite{arcade}. Each task has 1,000 training, 200 validation and 300 test images. The release is downloadable from Zenodo, but the archive metadata do not provide a simple redistribution license; we therefore redistribute scripts and manifests, not the image archive, and instruct users to obtain it under the stated terms.

Before modelling, we performed a read-only identity audit. Numeric identifiers did not define a valid cross-task pairing: the train and validation folders were pixel-mismatched, whereas the test split contained both exact duplicate pairs and mismatches. No syntax annotation was attached to a stenosis image. Images were resized to 128 by 128 pixels, and polygons were rasterized to binary masks. The release does not contain patient identifiers, so all estimates are image-level and no patient-level or treatment-outcome inference is made.

\subsection{Fixed task model}

The model is a small shared convolutional encoder with syntax and stenosis heads. Syntax images contribute only a vessel-support loss; stenosis images contribute only a lesion-mask loss. A projection penalty is used as an auditable representation probe:

\begin{equation}
\mathcal{L}=\mathcal{L}_{\mathrm{syntax}}+\mathcal{L}_{\mathrm{stenosis}}+\lambda(\hat u^{\mathsf T}\hat v)^2,
\end{equation}

where the normalized pooled projections \(\hat u\) and \(\hat v\) are trained with \(\lambda=0.05\). The matched ablation removes this term while keeping the architecture, optimizer, data, epochs and parameter count unchanged. The term is not interpreted as biological disentanglement.

\subsection{Verifier state, gate and actions}

For a candidate lesion mask \(C\) and a predicted vessel-support mask \(V\), the verifier state is

\begin{equation}
s=[\rho_V,\rho_C,o,c_x,c_y], \qquad o=\frac{|V_{\mathrm{core}}\cap C|}{|C|+\epsilon},
\end{equation}

where \(V_{\mathrm{core}}\) is a conservative erosion of \(V\), \(\rho\) are mask fractions and \((c_x,c_y)\) is the normalized candidate centroid. The fixed heuristic gate retains a candidate only when its area is at least 20 pixels, no more than 25\% of the image and \(o\geq0.25\). Otherwise it abstains. These three gate boundaries are implementation constants; only the prediction score threshold described below was selected on validation data. No independent calibration of the heuristic gate boundaries is claimed.

The action audit applies four deterministic transformations: retain, horizontal reflection, vertical shift and dilation. Each action produces a new verifier state that is recorded alongside the original state. These actions are controlled audit proxies. They do not represent catheter mechanics, device response or a clinical treatment transition.

\subsection{Evaluation and uncertainty}

The stenosis threshold was selected on validation data from 0.10 to 0.90 in increments of 0.05 by mean image-level Dice, then locked before test evaluation. We report image-bootstrap 95\% intervals from 1,000 resamples for syntax and stenosis Dice; AUROC, AUPRC, coverage and action acceptance are point estimates. Controlled shifts were horizontal reflection, an eight-pixel vertical shift and intensity attenuation. A separate constructed fault bank applied reflection, shift, random-frame candidate-mask substitution and dilation to reference masks while keeping the predicted support for the receiving frame fixed. Random substitution sampled a candidate mask from the test-frame pool rather than replacing an angiography image. These transformations were not independently adjudicated as incorrect lesions: transformed masks can retain valid overlap, and random sampling can select the original frame. Their acceptance is therefore an audit response, not an estimate of clinically adjudicated false acceptance. Reference masks provide a separate acceptance comparator. These are programmed audit environments, not external-cohort validation or a preregistered clinical trial.

\section{Results}

\subsection{Locked test performance}

The orthogonal model selected a stenosis threshold of 0.80 on validation. Locked test performance is shown in Table~\ref{tab:locked}. Syntax Dice was 0.475 (95\% CI 0.459--0.490), stenosis Dice was 0.231 (0.214--0.249), AUROC was 0.957 and AUPRC was 0.180 at a pixel prevalence of 0.00975. The matched no-orthogonality model selected 0.75 and achieved syntax Dice 0.457, stenosis Dice 0.236, AUROC 0.956 and AUPRC 0.186. The representation penalty therefore did not establish a consistent predictive advantage.

\begin{table}[t]
\centering
\caption{Locked test metrics. Values are image-level estimates from the public ARCADE test environments.}
\label{tab:locked}
\begin{tabular}{lccccc}
\toprule
Model & Threshold & Syntax Dice & Stenosis Dice & AUROC & AUPRC\\
\midrule
Orthogonal probe & 0.80 & 0.475 & 0.231 & 0.957 & 0.180\\
No orthogonality & 0.75 & 0.457 & 0.236 & 0.956 & 0.186\\
\bottomrule
\end{tabular}
\end{table}

\begin{figure}[t]
\centering
\includegraphics[width=0.98\linewidth]{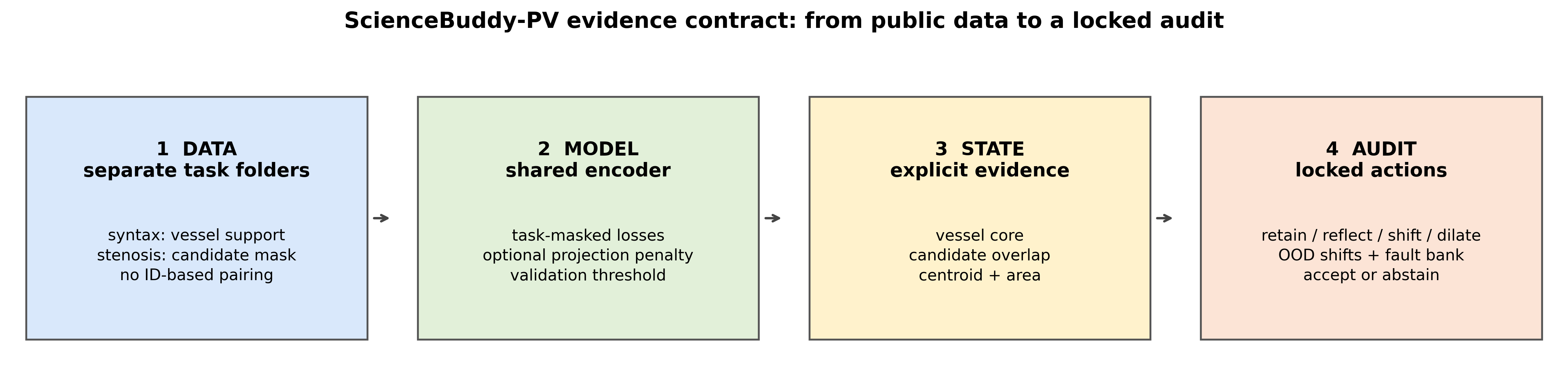}
\caption{ScienceBuddy-PV workflow. Separate public task environments feed task-masked training; a locked verifier exposes support and abstention; actions and fault injections form audit environments. No paired syntax and stenosis observation is assumed.}
\label{fig:workflow}
\end{figure}

\subsection{Support gate and action-state audit}

A score-only comparator emitted 100\% of stenosis-frame candidates. The evidence gate accepted 76.7\% at the validation-selected score threshold, with mean core-vessel overlap 0.311. Retain acceptance was 76.7\%, compared with 0\% after horizontal reflection, 0\% after vertical shift and 0.3\% after dilation. The gate therefore creates a visible abstention state and responds to constructed candidate transformations. These acceptance rates do not independently establish candidate correctness or clinical safety.

\begin{figure}[t]
\centering
\includegraphics[width=0.82\linewidth]{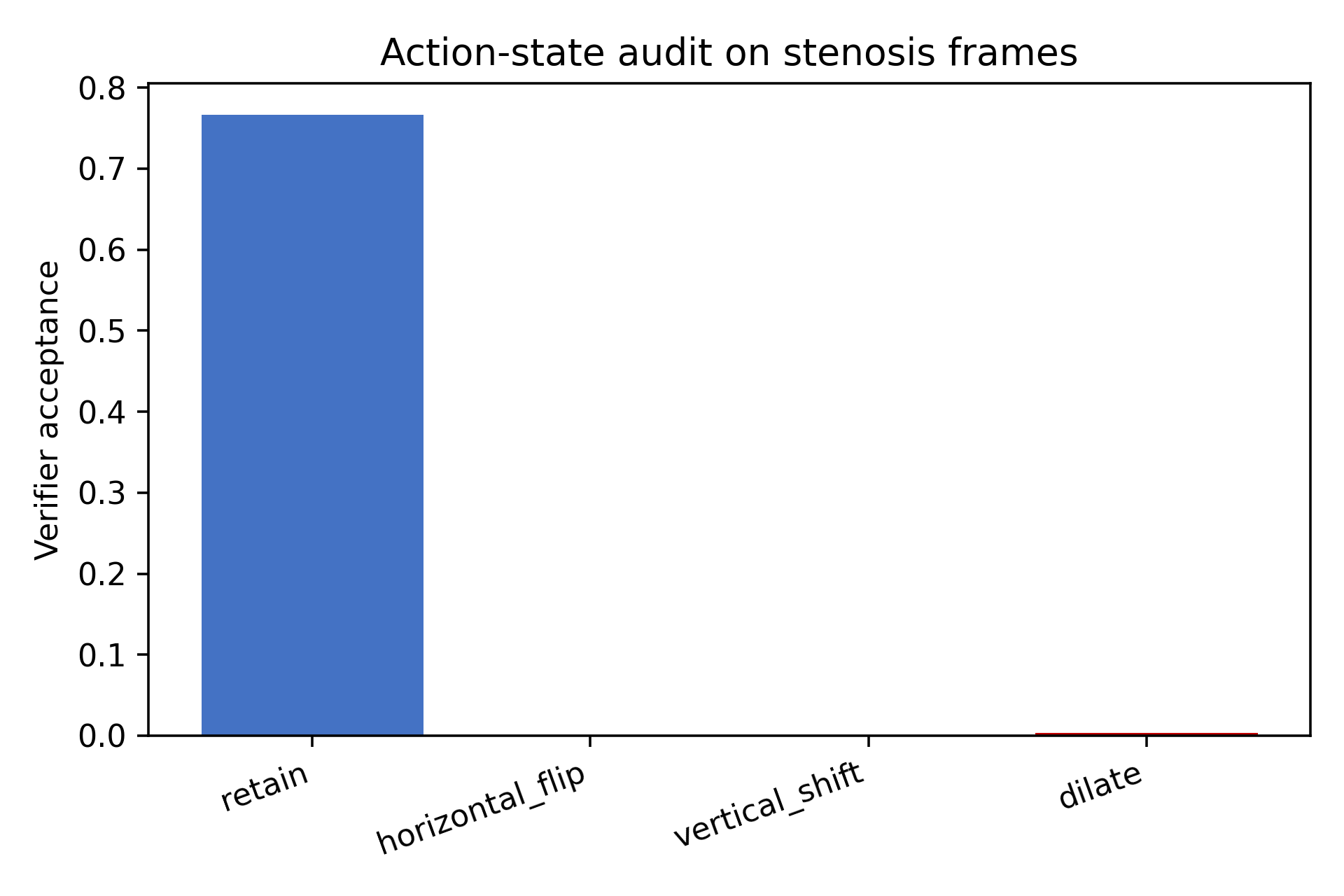}
\caption{Fixed-verifier action audit at the validation-selected score threshold. Constructed geometric transformations of predicted candidate masks alter the verifier state and usually trigger abstention; these are not learned treatment transitions.}
\label{fig:action}
\end{figure}

\subsection{Controlled shifts and fault injection}

Under horizontal reflection, vertical shift and intensity attenuation, stenosis Dice was 0.195, 0.228 and 0.170; AUROC was 0.940, 0.952 and 0.944; and AUPRC was 0.118, 0.175 and 0.131, respectively. Gate coverage was 0.633, 0.790 and 0.670. Coverage therefore cannot be interpreted as evidence that segmentation performance is preserved under a controlled shift.

In the constructed fault bank, reference candidate masks were accepted in 44.3\% of cases and rejected in 55.7\%. Reflection, shift, random-frame candidate-mask substitution and dilation were accepted in 2.0\%, 6.0\%, 3.0\% and 11.7\%, respectively. This comparison exposes the gate's rejection cost as well as its response to transformations. It does not show that every rejected candidate is unsupported or that every accepted candidate is correct.

\begin{figure}[t]
\centering
\includegraphics[width=0.86\linewidth]{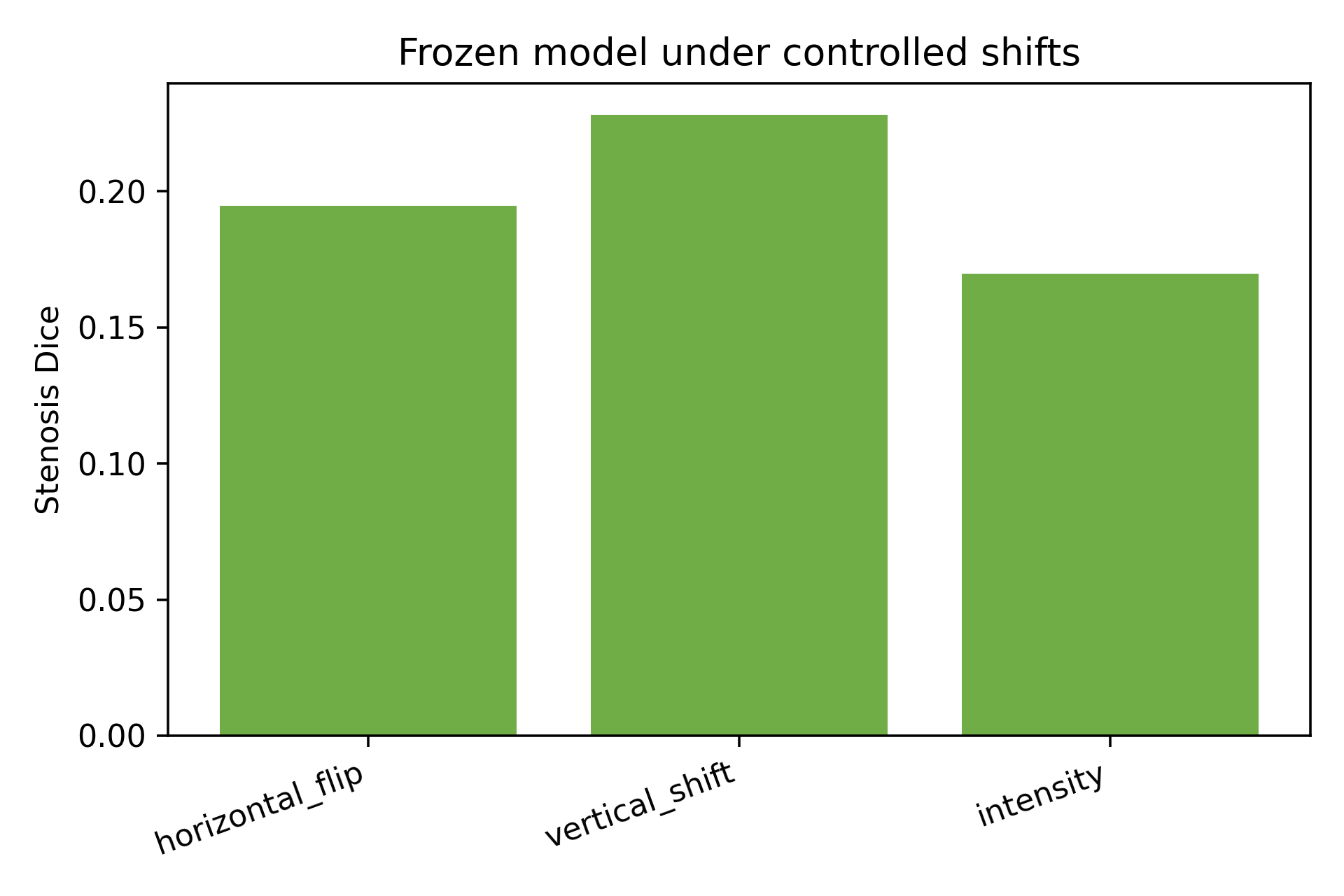}
\caption{Locked model under controlled distribution shifts.}
\label{fig:ood}
\end{figure}

\begin{figure}[t]
\centering
\includegraphics[width=0.90\linewidth]{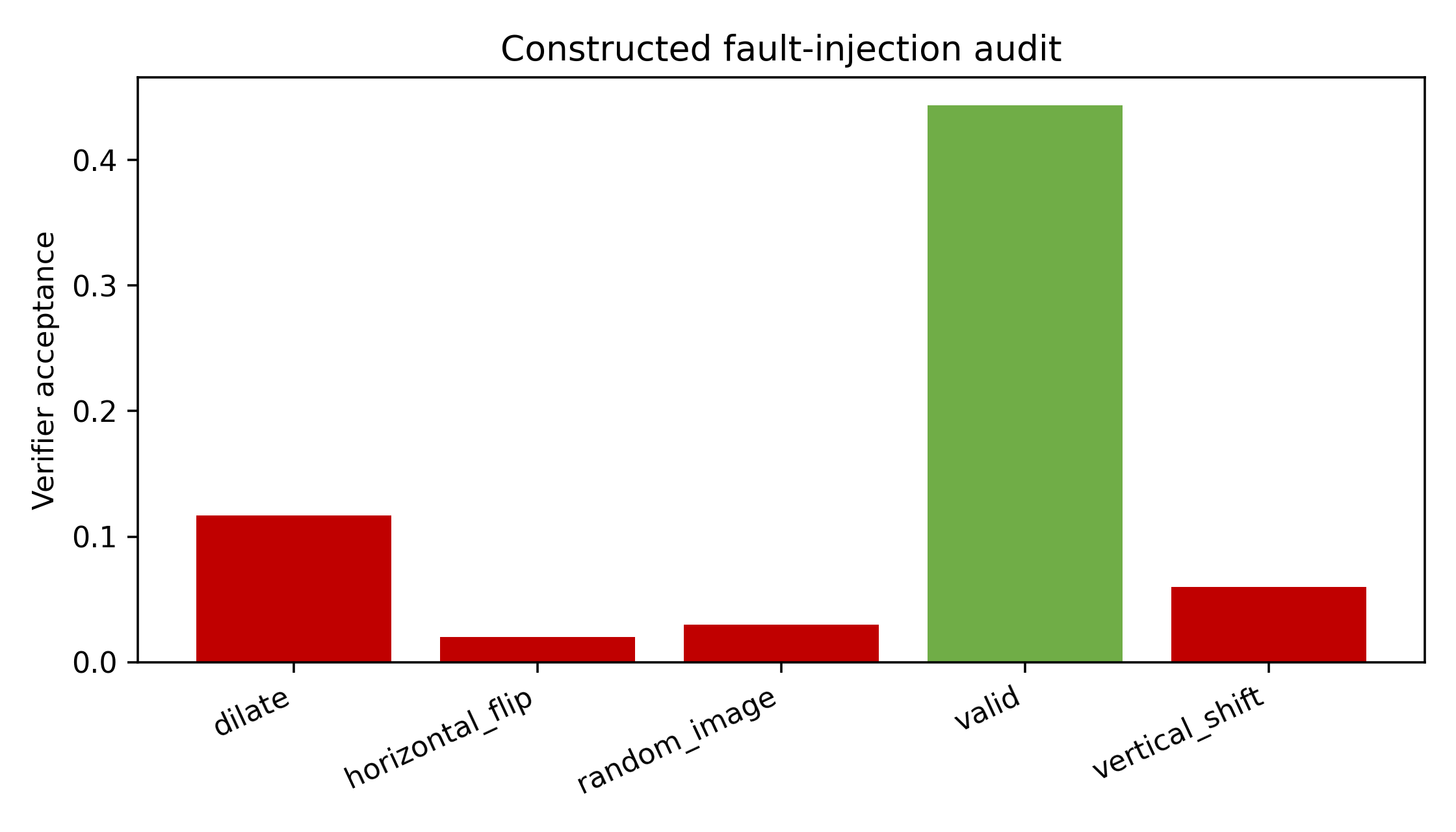}
\caption{Constructed fault-bank audit. Acceptance of reference and transformed candidate masks under the fixed support gate. Transformation labels are audit definitions rather than independently adjudicated errors.}
\label{fig:faults}
\end{figure}

\subsection{Expanded diagnostic panels}
The additional panels provide retrospective, descriptive views of the saved training histories, frozen test scores, validation threshold sweep, controlled shifts and fault bank. They introduce no new model selection or external cohort. Captions state the split and interpretation boundary so that these diagnostics cannot be mistaken for prospective or clinical validation.

\begin{figure}[t]
\centering
\includegraphics[width=0.98\linewidth]{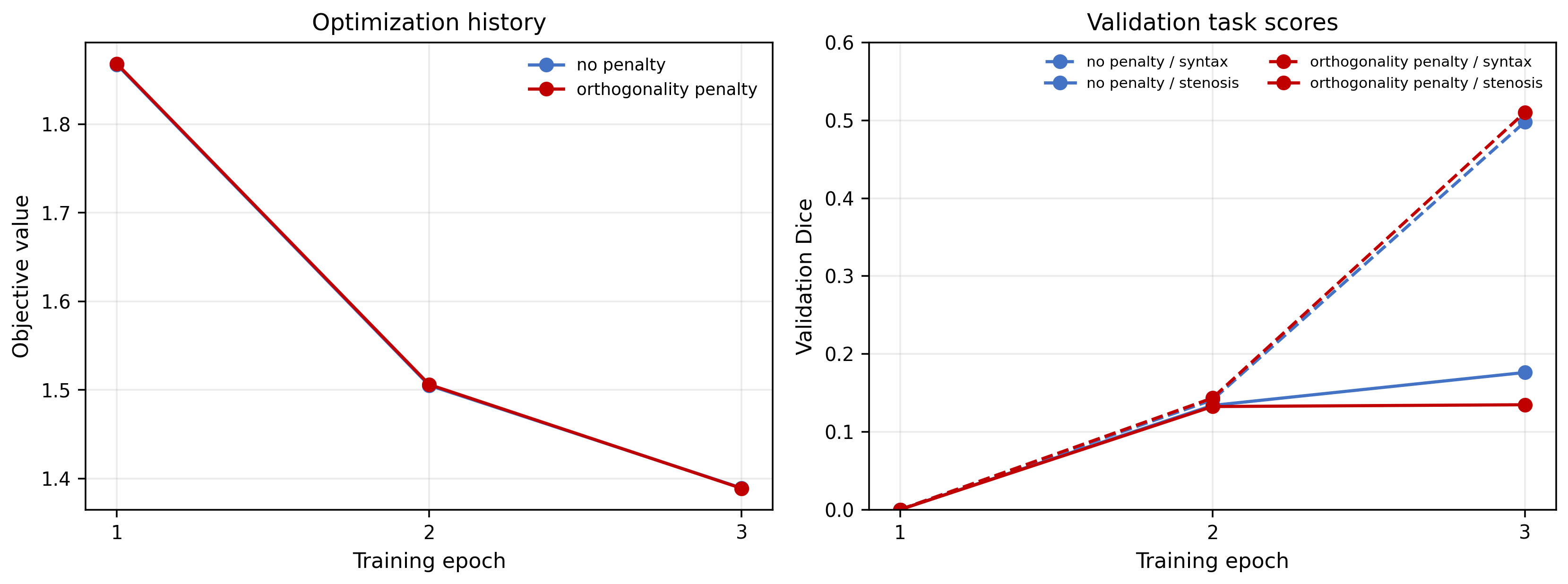}
\caption{Saved optimization histories for the matched representation comparison. Curves show objective value and validation Dice over three training epochs; the test model and threshold remained frozen for the locked audit.}
\label{fig:learning}
\end{figure}

\begin{figure}[t]
\centering
\includegraphics[width=0.92\linewidth]{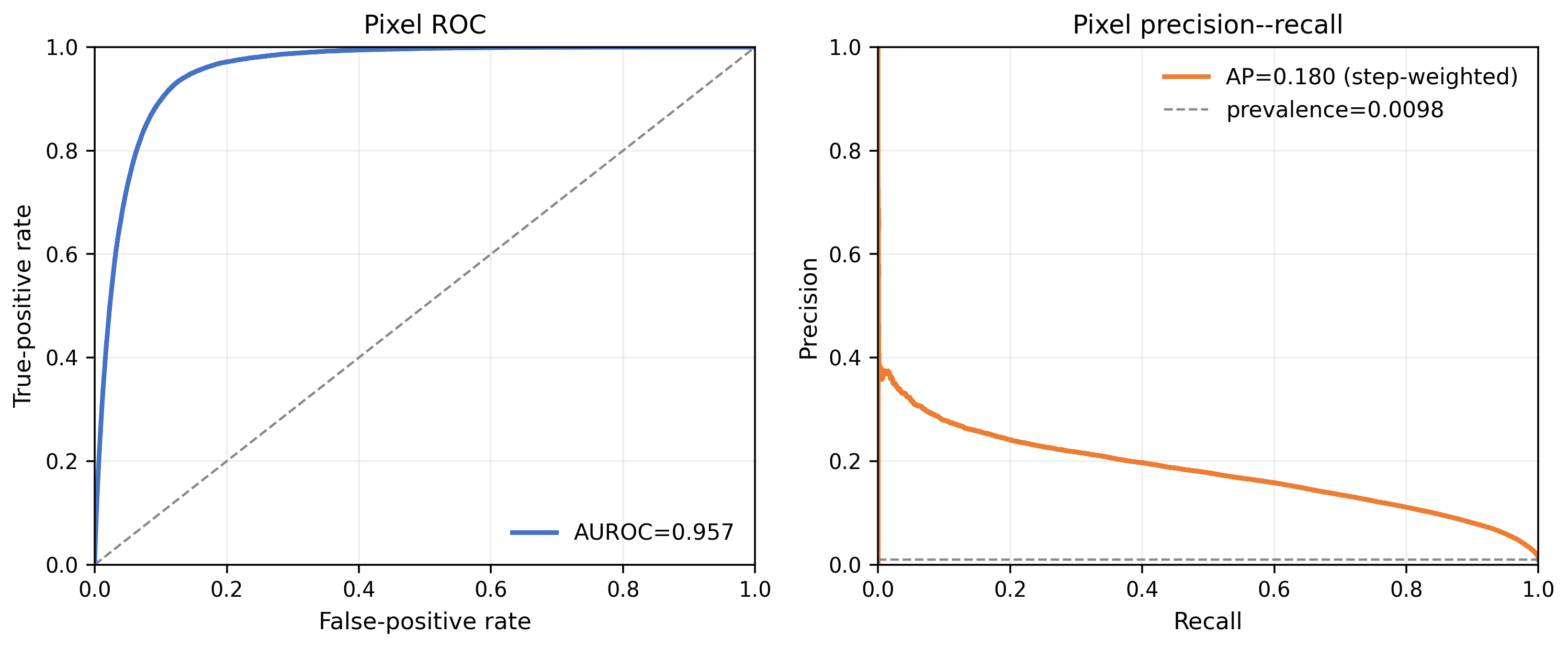}
\caption{Pixel-level discrimination on the untouched stenosis test environment. The precision--recall panel displays the low foreground prevalence explicitly; curves are computed from frozen model scores.}
\label{fig:pixel}
\end{figure}

\begin{figure}[t]
\centering
\includegraphics[width=0.96\linewidth]{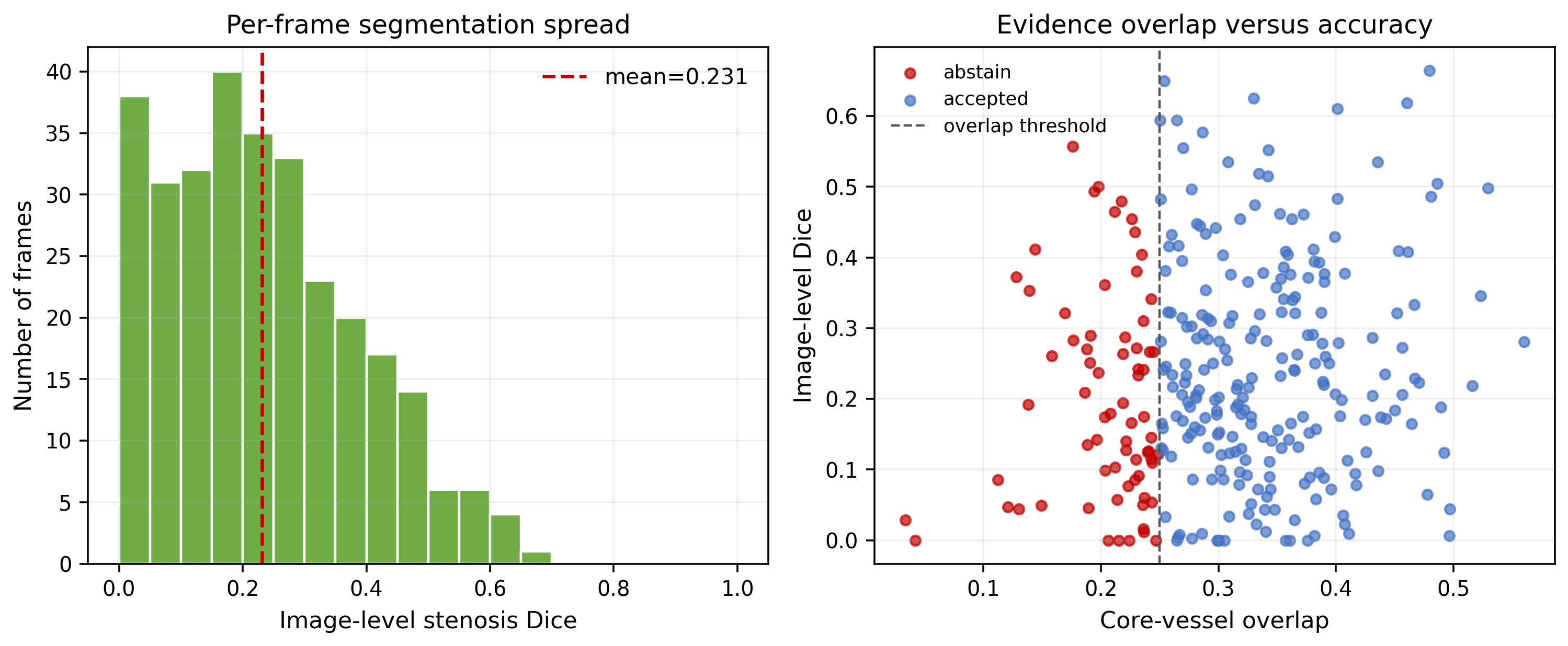}
\caption{Retrospective per-frame stenosis Dice and evidence overlap. Points are held-out frames; colors show whether the fixed heuristic core-vessel gate accepted or abstained at the validation-selected score threshold.}
\label{fig:stratification}
\end{figure}

\begin{figure}[t]
\centering
\includegraphics[width=0.98\linewidth]{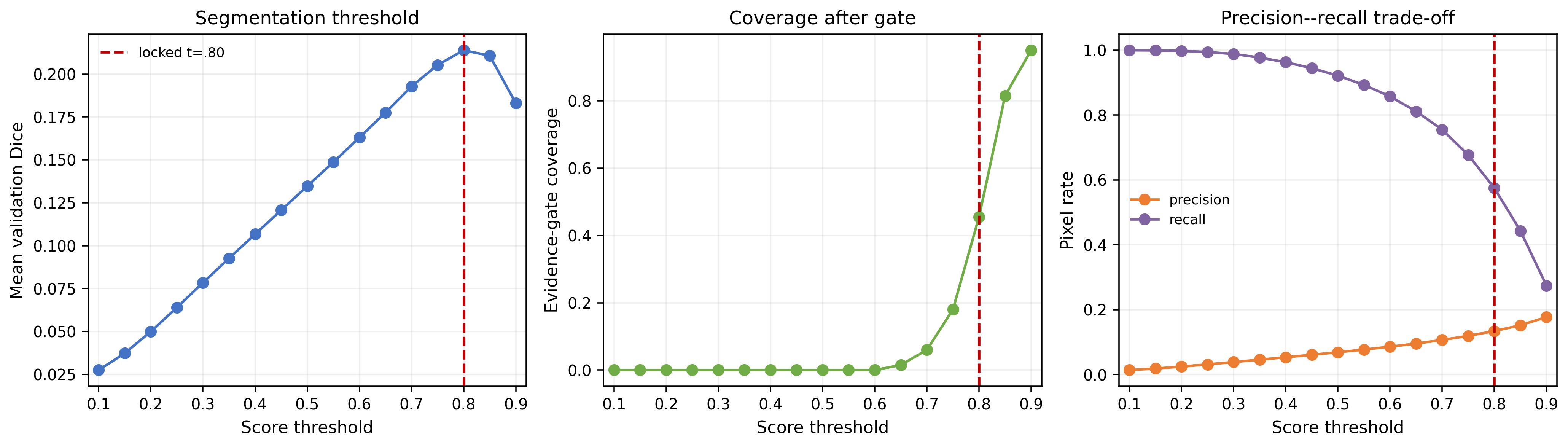}
\caption{Retrospective validation score-threshold diagnostic. The operating threshold (0.80) was selected on validation data before test evaluation; the panels show Dice, coverage of the fixed heuristic gate and pixel precision--recall.}
\label{fig:sensitivity}
\end{figure}

The learning curves (Figure~\ref{fig:learning}) show only a short three-epoch trajectory and do not establish convergence. Pixel discrimination (Figure~\ref{fig:pixel}) must be interpreted alongside the sparse foreground prevalence. The per-frame distribution and threshold sweep (Figures~\ref{fig:stratification} and \ref{fig:sensitivity}) expose heterogeneity that the pooled AUROC does not describe.

\begin{figure}[t]
\centering
\includegraphics[width=0.98\linewidth]{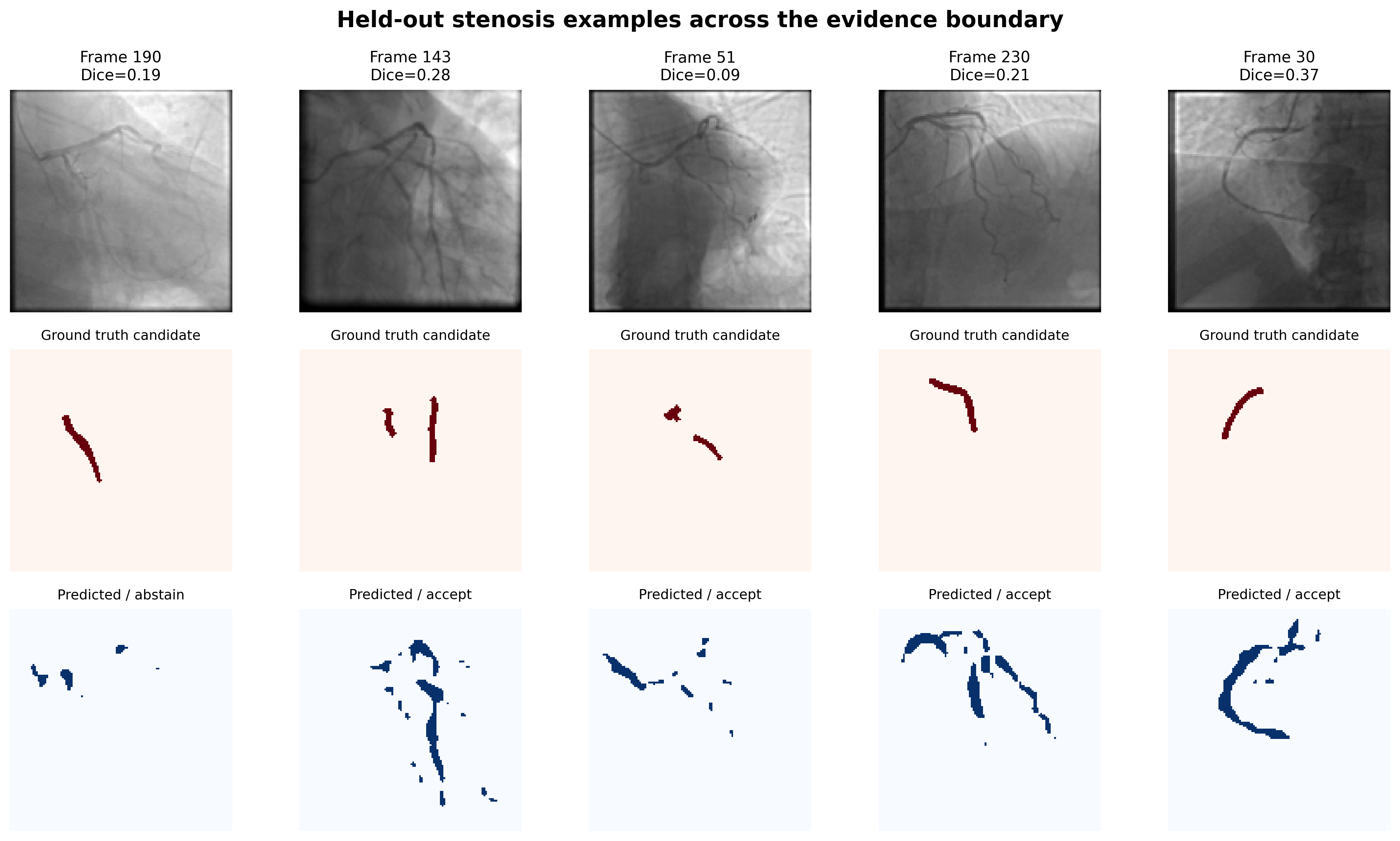}
\caption{Five retrospectively selected, deterministic held-out examples spanning the evidence boundary. Reference candidate masks are shown for technical comparison; acceptance is the fixed heuristic gate decision at the validation-selected score threshold.}
\label{fig:examples}
\end{figure}

\begin{figure}[t]
\centering
\includegraphics[width=0.98\linewidth]{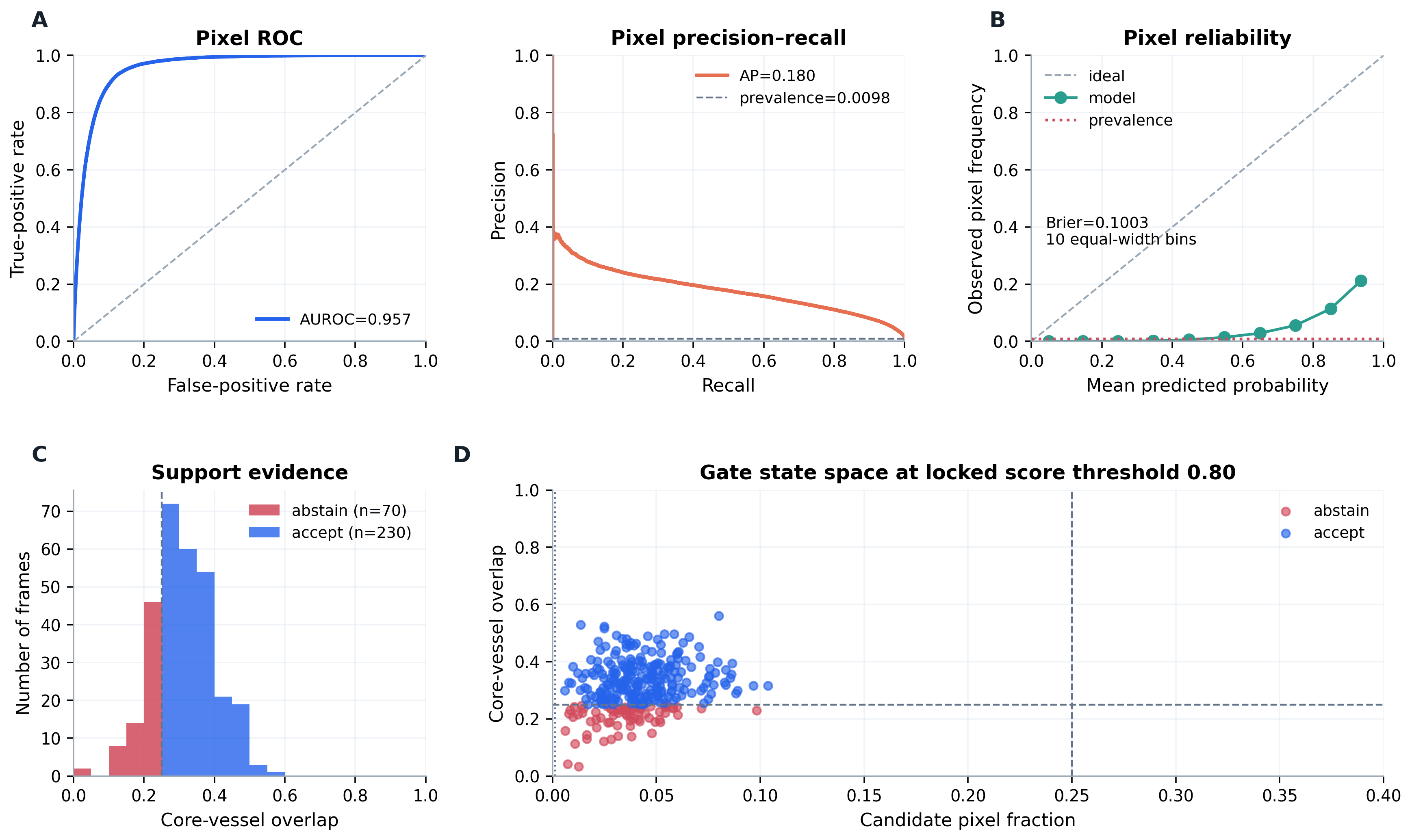}
\caption{Retrospective locked-evaluation diagnostics. (A) Pixel ROC and precision--recall curves on the untouched stenosis test environment. (B) Pixel-level reliability diagram with observed prevalence and Brier score. (C--D) Core-vessel overlap and candidate-area state space at the validation-selected prediction score threshold, with fixed heuristic gate boundaries. These are image-level technical quantities, not patient-level risk estimates.}
\label{fig:validation_panel}
\end{figure}

\begin{figure}[t]
\centering
\includegraphics[width=0.82\linewidth]{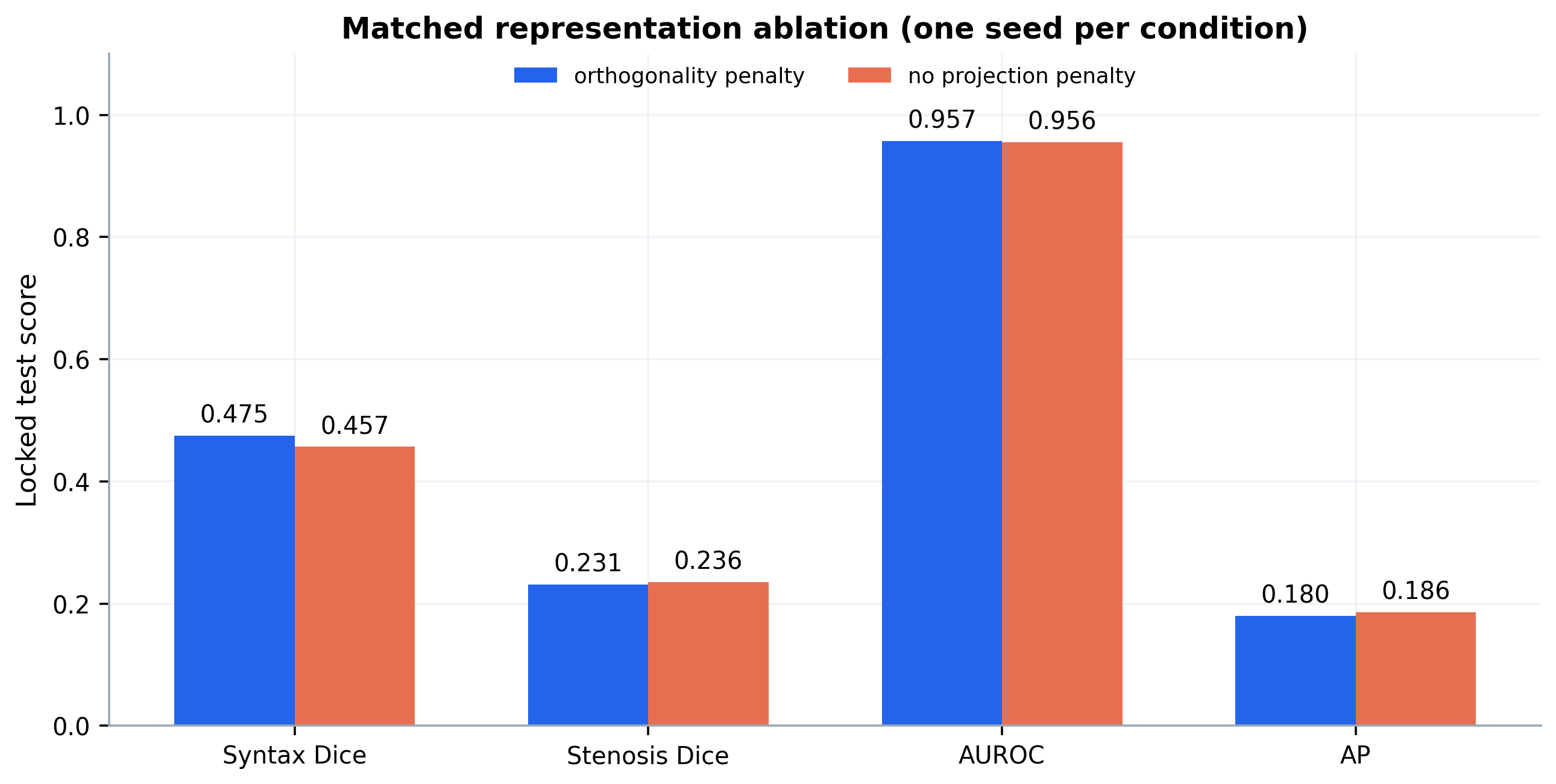}
\caption{Matched representation ablation. Locked test metrics for the orthogonality and no-penalty models; values are image-level estimates from the frozen ARCADE test environments.}
\label{fig:audit_ablation}
\end{figure}

\begin{figure}[t]
\centering
\includegraphics[width=0.98\linewidth]{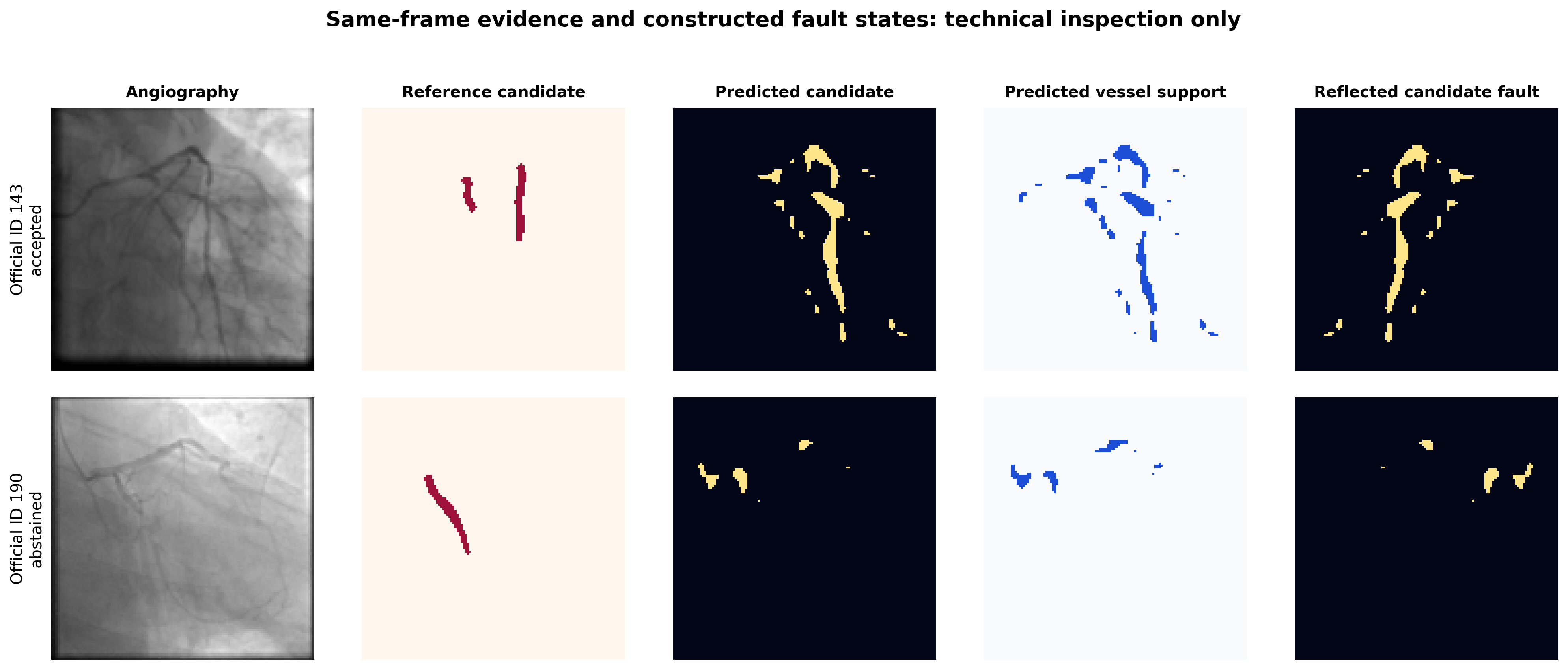}
\caption{Retrospective examples of evidence and constructed-fault states. Held-out angiography frames are paired with reference candidates, predicted candidates, predicted vessel support and reflected candidate-mask transformations for technical inspection only. Transformation labels are not independently adjudicated errors.}
\label{fig:audit_examples}
\end{figure}

The retrospective reliability audit (Figure~\ref{fig:validation_panel}B) yielded a pooled pixel Brier score of 0.1003. Predicted probabilities exceeded observed foreground frequencies across the ten bins, so the high ranking AUROC does not imply calibrated lesion probabilities. The state-space panels show that acceptance follows the specified geometric support rules; they are not evidence that accepted outputs are correct. The galleries (Figures~\ref{fig:examples} and \ref{fig:audit_examples}) preserve reference masks and constructed faults to make this limitation inspectable. The matched ablation (Figure~\ref{fig:audit_ablation}) retains the negative result that the projection penalty did not consistently improve prediction.

\begin{figure}[t]
\centering
\includegraphics[width=0.90\linewidth]{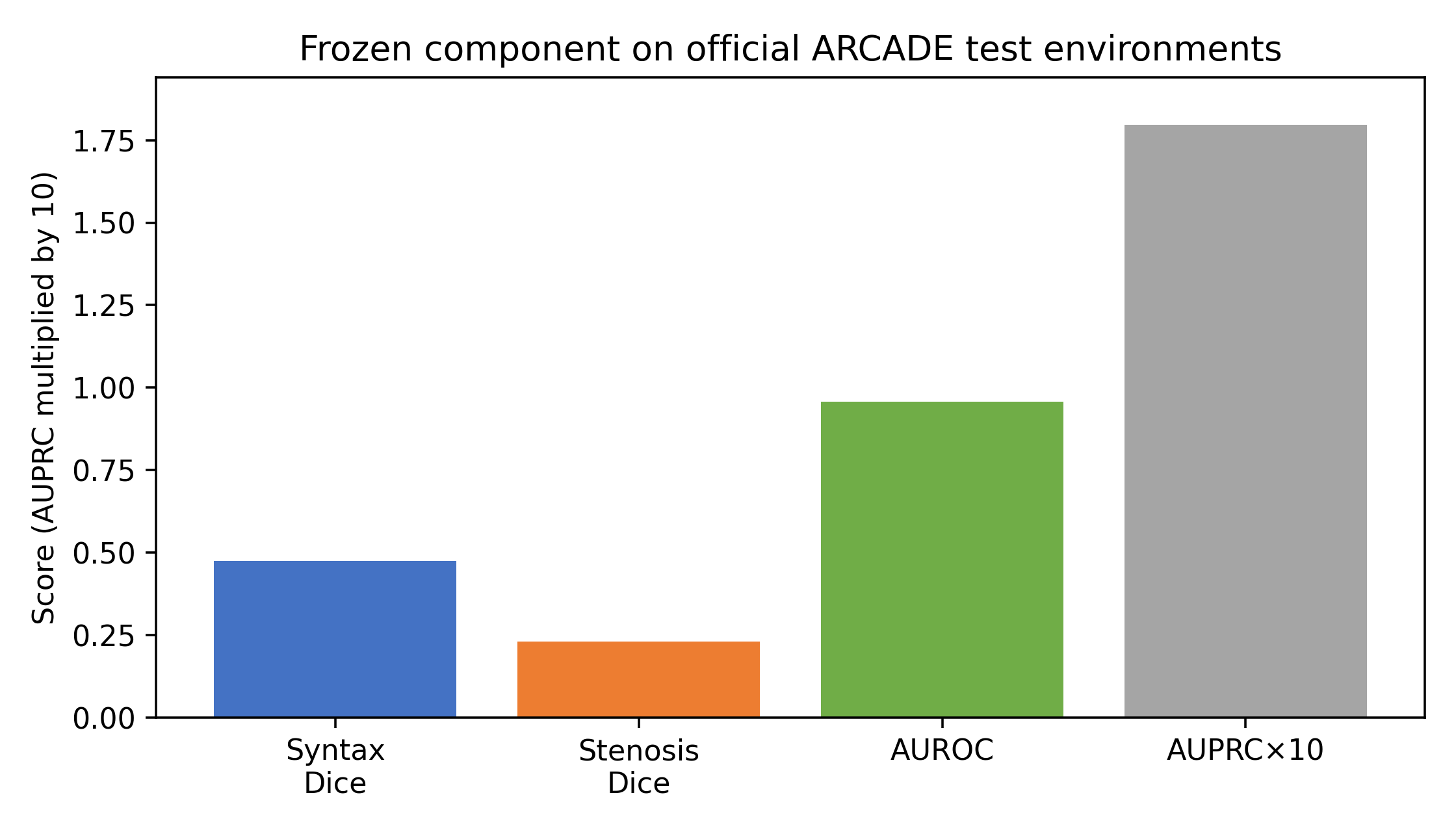}
\caption{Locked test metrics for the model trained with an orthogonality penalty. AUPRC is multiplied by 10 only for visual comparability; numerical values are reported in the text.}
\label{fig:metrics}
\end{figure}

\begin{figure}[t]
\centering
\includegraphics[width=0.78\linewidth]{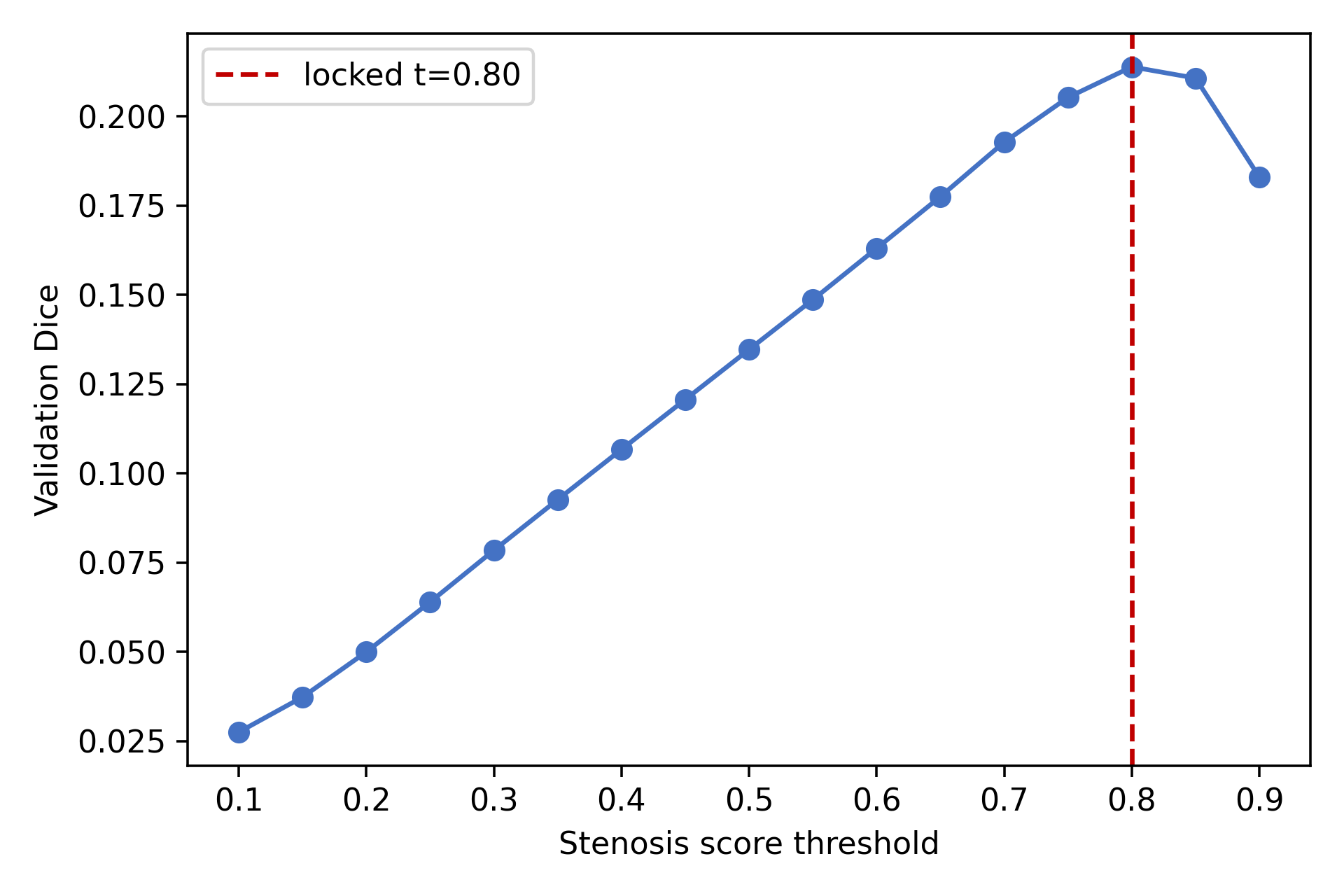}
\caption{Validation-only threshold selection. The dashed line marks the score threshold frozen before test evaluation.}
\label{fig:threshold}
\end{figure}

\begin{figure}[t]
\centering
\includegraphics[width=0.80\linewidth]{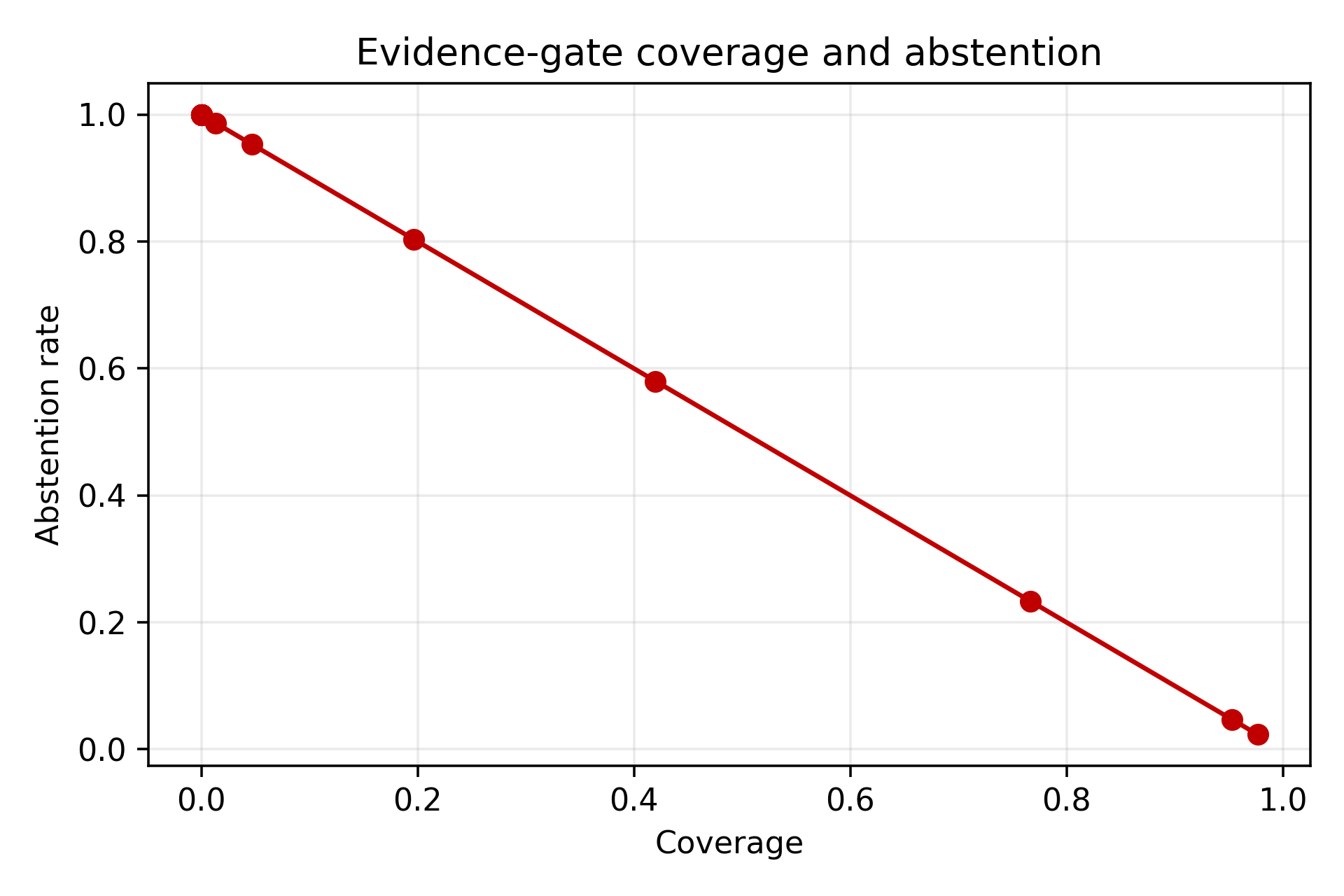}
\caption{Retrospective test-threshold diagnostic showing gate coverage and its complement, abstention. This identity does not estimate selective prediction risk. The reported operating score threshold was selected on validation data and then locked.}
\label{fig:coverage}
\end{figure}

\begin{figure}[t]
\centering
\includegraphics[width=0.95\linewidth]{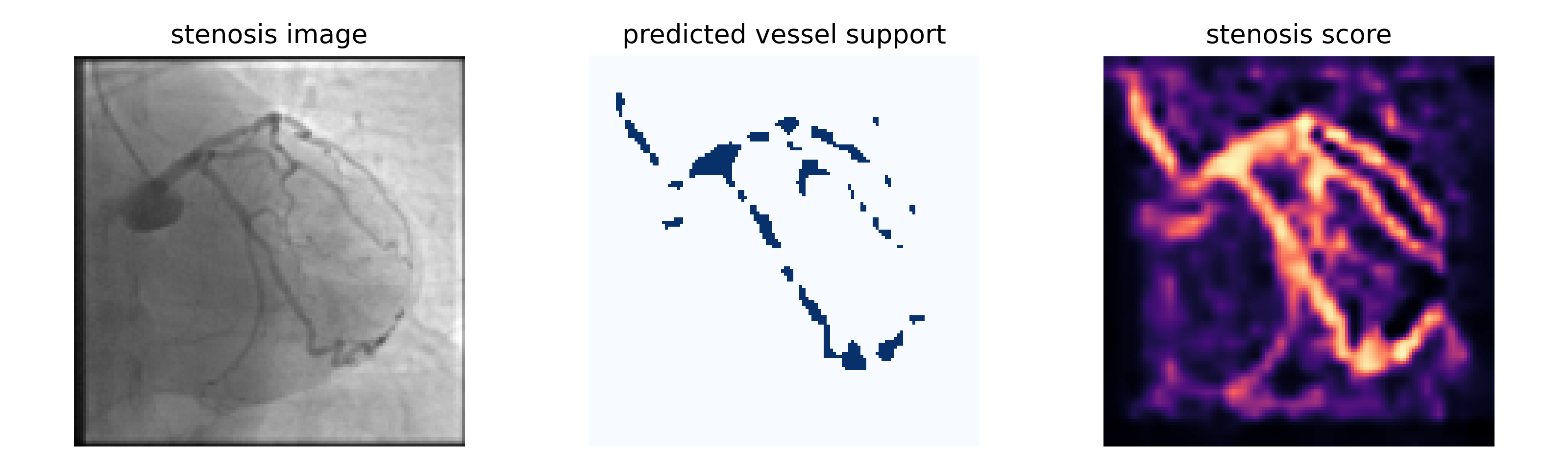}
\caption{Example held-out image, vessel-support prediction and stenosis score. These overlays are for technical inspection and are not clinical measurements.}
\label{fig:example}
\end{figure}

\begin{figure}[t]
\centering
\includegraphics[width=0.90\linewidth]{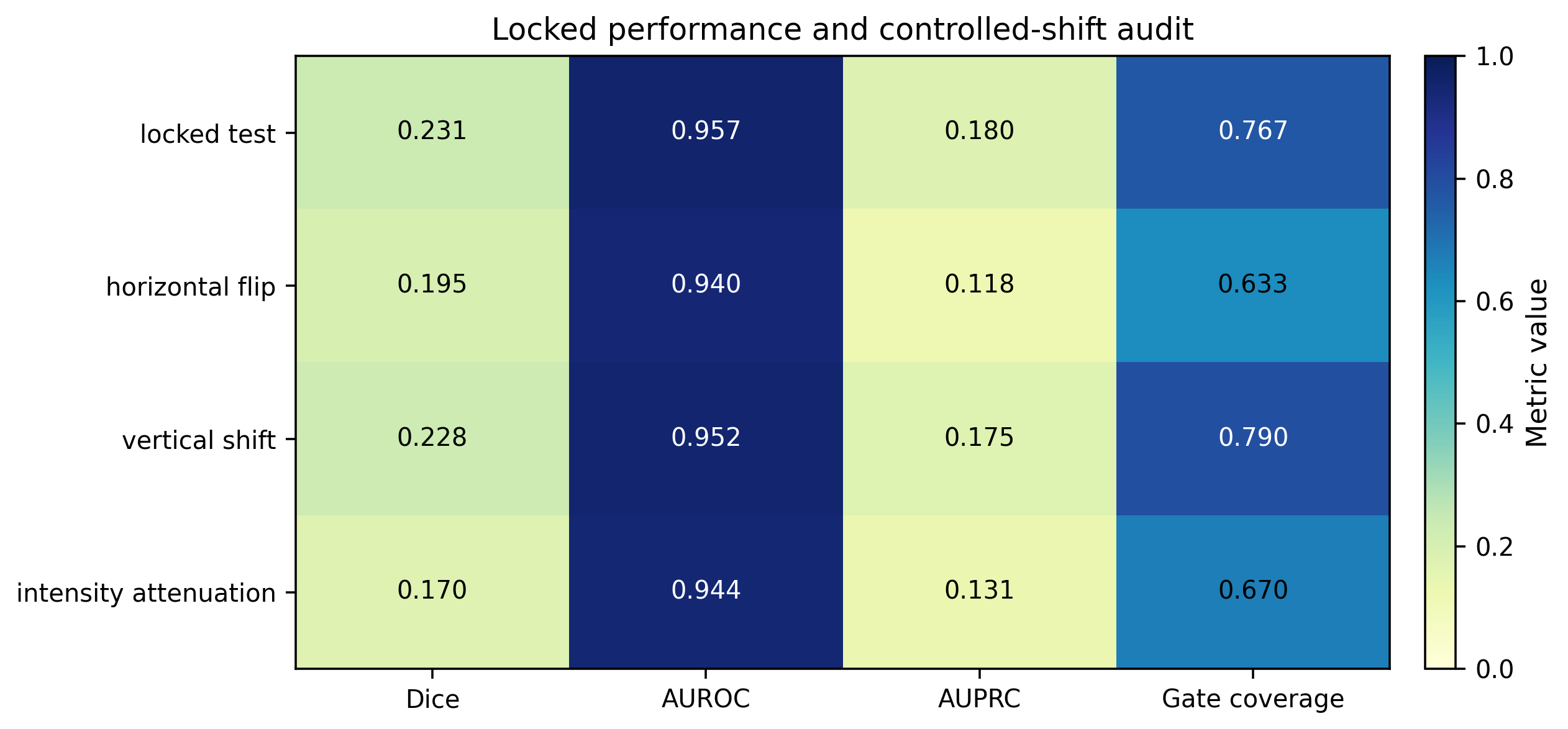}
\caption{Retrospective view of locked test metrics and the controlled-shift audit matrix. Values reproduce the saved summary; shifts are stress environments rather than external-cohort validation.}
\label{fig:shiftmatrix}
\end{figure}

\FloatBarrier
\section{Public-data extension for panvascular research}

The ARCADE experiment is intentionally narrow. A reproducible panvascular evidence agent can add public sources without pretending that they contain device outcomes:

\begin{itemize}[leftmargin=*]
\item PubMed and PMC Open Access supply source-level literature tasks and endpoint definitions.
\item ClinicalTrials.gov and AACT supply protocol, intervention, comparator and follow-up fields.
\item openFDA and MAUDE can support device-signal extraction, but their reports have no denominator and cannot establish incidence or comparative risk \cite{openfda}.
\item CADICA v5 offers a CC BY 4.0 coronary image benchmark with patient-level metadata; its role is image-to-geometry validation, not long-term device efficacy \cite{cadica}.
\item AortaSeg24 and open carotid-ultrasound datasets support geometry and segmentation probes without being relabelled as intervention outcomes \cite{aortaseg}.
\item MIMIC-IV and eICU are credentialed-access resources for future temporal or external-hospital analyses, not freely downloadable raw patient datasets \cite{mimic,eicu}.
\item PTB-XL can support future ECG signal tasks; its records are not used as intervention outcome labels \cite{ptbxl}.
\end{itemize}

The first agent task suite should contain literature reading, endpoint normalization, evidence conflict resolution, four-axis device--vessel matching and next-experiment planning. A minimum of 240 tasks can be separated by source identifier and evidence cutoff into development, internal test, temporal external and leave-one-bed-out sets. Each task should be independently reviewed by two vascular or interventional researchers, with a third adjudicator for conflicts. A rubric can score citation entailment, field completeness, units, reproducibility, uncertainty and safety. A fabricated citation, simulated result presented as clinical evidence, endpoint conflation or unsupported device recommendation is a key safety error and should cap the task score.

\section{Discussion}

This study proposes a vascular research protocol in which reviewer corrections can become executable checks, and tests a component that exposes its support state. The public benchmark shows three concrete behaviors. First, a score-only model can be wrapped in a gate that abstains on a measurable fraction of outputs. Second, spatial or intensity shifts can reduce lesion performance while leaving non-zero coverage, so coverage is not an assurance metric. Third, a constructed fault bank makes the gate's response to mask transformations visible. The present experiment does not establish recursive improvement, clinical correctness or a reduction in adjudicated scientific errors.

The acceptance of only 44.3\% of reference candidate masks is a substantial limitation: the gate rejects 55.7\% even when supplied with a reference lesion candidate. This behavior may reflect imperfect predicted vessel support, the area limits or the overlap heuristic. The present audit does not disentangle those causes. A future validation should independently adjudicate candidate validity and measure accepted-output quality, reference-candidate sensitivity and error-conditioned coverage before interpreting abstention as a safety benefit.

The experiment also clarifies what has not been demonstrated. ARCADE is image-level and does not contain patient identifiers, procedural actions, device specifications, blood flow, longitudinal outcomes or clinical adjudication. Its masks cannot establish Suitcordance, a digital twin or treatment benefit. The four Suitcordance axes are therefore an ontology for future linked studies, not a patient-level score in this paper. A new vascular bed must be recalibrated; a ranking that transfers across beds does not imply that an absolute risk transfers.

The outer recursion is the next research stage rather than a hidden claim of the current result. Its first safe implementation should use frozen evidence tasks, a human-reviewed rubric and a model update performed on public or synthetic trajectories. The updated model must pass the inherited harness, a blind red-team set and a temporal source holdout before any new interaction is collected for learning. Positive reward for correct refusal is especially important in this domain because a longer answer can otherwise look more useful while increasing unsupported device claims.

The most informative next project is a narrow, bed-specific study of device--vessel concordance. Coronary data should be used for development, with femoropopliteal data as a deliberately difficult external transfer. Primary endpoints must remain bed-specific and source-defined; CD-TLR, TLF, restenosis, primary patency, limb events and death should not be averaged into a generic success label. Geometry, mechanics, haemodynamics and biology should be measured separately before any composite is considered. CFD/FEM and FLOW-style experiments can generate controlled state--action--outcome pairs, but they must enter as soft constraints or falsification benchmarks rather than as substitutes for clinical follow-up.

\section{Limitations and safety boundaries}

The benchmark is limited by public image-level labels, an incomplete clinical ontology and controlled rather than real-world shifts. Each representation setting was trained for three epochs with one random seed; training variability and convergence are therefore not characterized. The projection penalty does not prove disentanglement. The verifier boundaries are fixed heuristics and may not be appropriate for other tasks. Constructed fault labels do not establish true error status, and the gate rejects more than half of reference candidates. The expanded diagnostics are retrospective. Recursive harness optimization, researcher feedback experiments and outer model learning have not been executed. No output should be used to select a device for an individual patient. Any future clinical extension requires governance approval, de-identification, role-based access, immutable audit logs, out-of-distribution monitoring, human sign-off and prospective silent validation.

\section{Reproducibility and data availability}

The source package contains training, evaluation, threshold-selection and audit scripts, results JSON files, split manifests and the manuscript figures. The ARCADE image archive is not redistributed. Users should obtain it from the public Zenodo record under its stated access terms, run the preparation script, then execute the training and audit scripts. The saved results distinguish the primary validation-selected metrics from the uncalibrated training-stage diagnostics. All 17 figure items have a generation path documented in the ancillary figure-provenance manifest. The added panels are retrospective diagnostics from frozen checkpoints and saved summaries; the included checkpoints permit their regeneration after the public images are prepared, without retraining. The package records dependency versions and per-figure hashes. The package also contains a public-data plan for CADICA, PTB-XL, AACT, ClinicalTrials.gov, openFDA, AortaSeg24 and credentialed-access resources; those datasets are not mixed into the primary ARCADE result.

\section*{Declaration of generative AI}

Codex assisted with literature retrieval, LaTeX assembly, code organization and consistency checks. The authors are responsible for source verification, data governance, analysis decisions and the final manuscript.


\begin{thebibliography}{19}
\bibitem{sciencebuddy} Xue S, Zhong J, Nan Z, et al. ScienceBuddy: Recursive-in-Recursive Self-Improvement for Interactive Scientific Agents. arXiv:2609.17523 (2026). \url{https://arxiv.org/abs/2609.17523}.
\bibitem{you_suitcordance} You L, Chen Y, Zhang Z, et al. High Suitcordance for Panvascular Full-Watershed Organs: A New Interventional Perspective. Research. 2025;8:0974. doi:10.34133/research.0974.
\bibitem{suitbench} You L. Suitcordance Bench: device--vessel matching across the arterial tree. Research prototype. \url{https://lingsenyou.com/suitcordance/}.
\bibitem{you_radar} You L. Lingsen You: Interventional Cardiology and Panvascular Research. \url{https://lingsenyou.com/}.
\bibitem{flow} You L, et al. The FLOW Framework: a panvascular-on-a-chip platform to model systemic disease and guide panvascular interventional device suitcordance. Science Bulletin. 2026;71(3):654--670. doi:10.1016/j.scib.2025.12.051.
\bibitem{4a} You L, et al. 4A tetrahedron system: a synergistic framework for panvascular intervention empowered by flexible electronics. npj Flexible Electronics. 2026;10:35. doi:10.1038/s41528-026-00537-5.
\bibitem{digitaltwin} Skalidis I, Stalikas N, Collet C, et al. Digital twins and simulations in transcatheter coronary and structural heart interventions. Eur Heart J Digit Health. 2026;7(2):ztaf129. doi:10.1093/ehjdh/ztaf129.
\bibitem{arcade} Popov M, Amanturdieva A, Zhaksylyk N, et al. Dataset for Automatic Region-based Coronary Artery Disease Diagnostics Using X-Ray Angiography Images. Sci Data. 2024;11:20. doi:10.1038/s41597-023-02871-z.
\bibitem{jepa} Cui T, Wang Z, Xu X, et al. JEPA-Anything: Learning Predictive Models across Different Worlds. arXiv:2609.20800 (2026). \url{https://arxiv.org/abs/2609.20800}.
\bibitem{iwm} Cui T, Wang X, Li Z, et al. An immune world model for multiscale forecasting and therapeutic hypothesis generation. arXiv:2609.14709 (2026). \url{https://arxiv.org/abs/2609.14709}.
\bibitem{scienceide} Geng H, Huang Z, Li H, et al. ScienceIDE: Turning World's Scientific Codebase into Agent Learnable Environments. arXiv:2609.19134 (2026). \url{https://arxiv.org/abs/2609.19134}.
\bibitem{samant} Samant S, Panagopoulos AN, Wu W, Zhao S, Chatzizisis YS. Artificial Intelligence in Coronary Artery Interventions: Preprocedural Planning and Procedural Assistance. J Soc Cardiovasc Angiogr Interv. 2025;4(3 Pt B):102519. doi:10.1016/j.jscai.2024.102519.
\bibitem{ivus} Matsumura M, Mintz GS, Dohi T, et al. Accuracy of IVUS-Based Machine Learning Segmentation Assessment of Coronary Artery Dimensions and Balloon Sizing. JACC Adv. 2023;2(7):100564. doi:10.1016/j.jacadv.2023.100564.
\bibitem{cadica} Jimenez-Partinen A, Molina-Cabello MA, Thurnhofer-Hemsi K, et al. CADICA: a new dataset for coronary artery disease. Mendeley Data. Version 5. 2025. doi:10.17632/p9bpx9ctcv.5.
\bibitem{ptbxl} Wagner P, Strodthoff N, Bousseljot RD, et al. PTB-XL, a large publicly available electrocardiography dataset. Sci Data. 2020;7:154. doi:10.1038/s41597-020-0495-6.
\bibitem{mimic} Johnson AEW, Bulgarelli L, Shen L, et al. MIMIC-IV, a freely accessible electronic health record dataset. Sci Data. 2023;10:1. doi:10.1038/s41597-022-01899-x.
\bibitem{eicu} Pollard TJ, Johnson AEW, Raffa JD, et al. The eICU Collaborative Research Database, a freely available multi-center database for critical care research. Sci Data. 2018;5:180178. doi:10.1038/sdata.2018.178.
\bibitem{openfda} U.S. Food and Drug Administration. openFDA device adverse event data. \url{https://open.fda.gov/apis/device/event/}.
\bibitem{aortaseg} Imran M, et al. Multi-class segmentation of aortic branches and zones in computed tomography angiography: The AortaSeg24 challenge. Med Image Anal. 2026;113:104188. doi:10.1016/j.media.2026.104188.
\end{thebibliography}
\end{document}